# Evaluative Fingerprints

*Stable and Systematic Differences in LLM Evaluator Behavior*

Wajid Nasser (Viore) • January 2026

**Abstract**

LLM-as-judge systems promise scalable, consistent evaluation. We find the opposite: judges are *consistent*, but not with each other; they are consistent with *themselves*.

Across 3,240 evaluations (9 judges × 120 unique video×pack items × 3 independent runs), inter-judge agreement is near-zero (Krippendorff's α = 0.042). On two dimensions, judges disagree more than random noise would predict (α < 0). Yet this disagreement isn't chaos; it's structured. A classifier identifies which judge produced an evaluation with 77.1% accuracy from rubric scores alone, rising to 89.9% with disposition features. Within model families, the signal is even stronger: GPT-4.1 and GPT-5.2 are distinguishable with 99.6% accuracy.

We call this the *reliability paradox*: judges cannot agree on what constitutes quality, yet their disagreement patterns are so stable they function as fingerprints. Each judge implements a distinct, stable theory of quality: an "evaluative disposition" that shapes how it interprets any rubric. We characterize these dispositions along multiple axes: harshness/leniency, dimension emphasis, within-judge stability (ICC), and evidence behavior (receipt validity, semantic linkage via NLI, and shotgun index).

The implication is stark: LLM judges are not interchangeable instruments measuring a shared construct. They are distinct measurement devices, each encoding its own implicit theory of quality. Averaging their scores produces a synthetic verdict that corresponds to no judge's actual values.

## 1. Introduction: The Uncomfortable Question

LLM-as-judge has become infrastructure. Benchmarks use it. RLHF pipelines use it. Product teams use it to score generations, rank candidates, and make shipping decisions. The assumption, usually implicit, is that these judges measure something real: that scores reflect quality, and different judges are noisy measurements of the same underlying truth.

But what if they're not?

What if each judge is measuring something different: not noise around a shared signal, but fundamentally incompatible theories of what "good" means?

This paper treats judges as measurement instruments and tests that uncomfortable hypothesis. We ask three questions:

1. **Agreement:** When multiple LLM judges evaluate the same content under the same rubric, do they agree?

2. **Stability:** Are individual judges consistent with themselves across repeated evaluations?

3. **Identifiability:** If judges disagree systematically, can we identify which judge produced an evaluation from its scores alone?

The answers reveal a paradox. Judges barely agree with each other (Question 1), yet many are highly consistent with themselves (Question 2), and their patterns are distinctive enough to identify them (Question 3).

**In other words: judges can't agree on what's good, yet they're so consistent in how they disagree that you can identify them.**

This has practical consequences. If you run your benchmark with Claude instead of GPT, you might declare a different winner. If you average judges to get a "consensus," you get a synthetic score that matches no judge's actual assessment. Model choice isn't an implementation detail; it's a substantive methodological decision that shapes what you're measuring.

## 2. Related Work

### 2.1 LLM-as-Judge: Foundations and Known Biases

MT-Bench and Chatbot Arena established LLM judging as a scalable alternative to human evaluation (Zheng et al., 2023). G-Eval demonstrated rubric-driven prompting with GPT-4 (Liu et al., 2023). Subsequent work documented systematic biases: position effects, verbosity preferences, self-enhancement, and sensitivity to prompt formatting (Wang et al., 2024; Gu et al., 2024).

Recent work has also questioned the reliability of LLM judges across repeated runs. Haldar and Hockenmaier (2025) quantify self-inconsistency in LLM-as-a-judge settings, showing that within-judge reliability varies substantially across models, tasks, and label scales. Related analyses of LLM evaluators report anchoring and familiarity biases and sensitivity to prompt variants (Stureborg et al., 2024). Our study uses a different domain, rubric, and protocol, but likewise finds large judge-to-judge variation in self-consistency; crucially, we show these differences are stable enough to enable attribution.

Beyond reliability, several works highlight threats to construct validity. Ye et al. (2024) identify a taxonomy of potential biases in LLM-as-a-judge and propose CALM, a principle-guided perturbation framework for quantifying them. Chehbouni et al. (2025) argue, using measurement theory, that the field often assumes validity and reliability properties that LLM judges may not satisfy. Complementary evidence comes from studies of fine-tuned judge models, which can perform well in-domain yet fail to generalize or preserve fairness across settings (Huang et al., 2025).

Much of this literature frames bias and inconsistency as error: deviations from a hypothetical true score to be mitigated. Our framing differs: we treat systematic preference, calibration, and even stochasticity as dispositions (stable properties of the judge) that can be measured and identified.

## 2.2 Model Fingerprinting and Attribution

Recent work shows that LLMs can be identified from their generated text. Hide-and-Seek demonstrates behavioral fingerprints (Iourovitski et al., 2024); stylistic fingerprinting achieves high attribution accuracy from writing patterns (Bitton et al., 2025). More broadly, Behavioral Fingerprinting uses diagnostic prompt suites to characterize interactive behavior across models (Pei et al., 2025).

These approaches fingerprint via *generation*: the content models produce. We fingerprint via *evaluation*: how models score content they didn't create. This is a harder test: the judge is constrained by a rubric and presented with fixed artifacts. If fingerprints still emerge, they reflect deep model characteristics, not surface generation style.

## 2.3 Our Contribution

We study evaluation behavior as a fingerprint. Specifically, we: (1) characterize a reliability paradox: near-zero inter-judge agreement alongside stable, structured within-judge behavior; (2) show that evaluation outputs (rubric scores, disposition features, and evidence behavior) enable robust attribution at the family, model, and version level, including within-provider discrimination; and (3) validate persistence under input perturbations and across a second content regime, with controls indicating the signal is not reducible to simple score-scale usage. Prior LLM-as-a-judge work primarily treats bias and reliability as limitations, while prior fingerprinting work targets generation; we connect these lines by demonstrating that the evaluator itself can be fingerprinted from its judgments.

## 3. Experimental Design

### 3.1 Dataset

We evaluate 30 YouTube videos spanning 15 topic categories (comedy, AI/ML, travel, sports, tech reviews, etc.). For each video, we generate 4 SEO content packs using diverse LLM generators (GPT-5.2, GPT-4.1, Gemini-3-Pro, and a fourth slot that alternates between Mistral-Large and Claude-Opus across videos). Thus each video has four packs, but across the corpus there are five possible pack IDs (3, 41, 45, 53, 523). Each pack is evaluated by 9 judges across 3 independent runs.

We restrict analysis to the intersection set: **120 unique video×pack pairs**, each evaluated 3 times, yielding **360 evaluation instances per judge**. With 9 judges, total evaluations = 3,240.

### 3.2 Judges

Nine frontier models spanning five high-level provider families:

| Family | Models |
|---|---|
| Anthropic | Claude-Opus-4.5, Claude-Sonnet-4.5 |
| OpenAI | GPT-5.2, GPT-4.1 |
| Google | Gemini-3-Pro-Preview |
| xAI | Grok-3 |
| Open-weights | DeepSeek-R1, Llama-405B, Mistral-Large |

In addition to the high-level family grouping above, some analyses report a finer 7-way provider lineage split: Anthropic, OpenAI, Google, xAI, DeepSeek, Meta/Llama, and Mistral.

### 3.3 Evaluation Protocol

Each judge receives the same prompt with a 5-dimension rubric: Intent & Angle, Coverage & Completeness, Faithfulness & Receipts, Readability & Structure, and SEO Mechanics. Judges output structured JSON with dimension scores (1-5), an overall score, and "receipts": quoted spans from source material supporting their assessments.

**Strict compliance:** We require valid JSON with no repairs or transformations. Request retries were rare (~1.5-2% of calls), dominated by malformed JSON outputs or provider rate limiting; we do not repair outputs and only analyze parseable, protocol-compliant evaluations. Models below 98% compliance were excluded (Cohere, Kimi K2).

## 4. Metrics

### 4.1 Agreement Metrics

**Rank agreement:** Pairwise Spearman rank correlation on item-level overall scores (run-averaged).

**Absolute agreement:** Krippendorff's α (interval) on the 9×360 rater×item matrix, computed per dimension and overall.

**Within-judge stability:** ICC(3,1) per judge across the three runs (items = 120 unique video×pack pairs).

### 4.2 Disposition Metrics

**Harshness:** For each evaluation row, harshness is the deviation from the across-judge mean for the same (video, pack, run). We report per-judge mean harshness with 95% video-cluster bootstrap percentile intervals (resampling videos; n_boot=2000).

**Dimension emphasis:** Per-judge harshness computed separately for each rubric dimension, revealing which aspects each judge weights more heavily.

### 4.3 Evidence Behavior: Provenance vs. Semantic Linkage

We analyze judge evidence behavior through a two-stage pipeline:

**Provenance (Presence Validity):** Fraction of receipts whose quoted evidence is found in the declared source text via normalization + fuzzy matching (threshold = 0.90). This verifies the judge actually cited real content.

Semantic Linkage (NLI): Conditional on presence-valid receipts, we test whether the cited quote can certify the judge's aggregated justification using Natural Language Inference (a DeBERTa-v3 NLI model fine-tuned on MNLI). This is a strict linkage/certification test, not a factual correctness test; because justifications are aggregated summaries while receipts are atomic spans, the linkage rate is a conservative lower bound. A receipt is marked "supported" if: (a) entailment probability ≥ 0.75, and (b) entailment margin (p_entail - p_contradict) ≥ 0.20. Justifications are truncated to the first 200 characters.

Human audit: Semantic Linkage (pilot).

We manually audited a 200-row, balanced pilot (100 predicted SUPPORTED, 100 predicted NOT_SUPPORTED). We used a strict certification criterion: the atomic receipt quote must certify the aggregated justification.

Agreement was 57.5% (115/200), with a false-positive skew: when NLI predicted SUPPORTED, humans agreed 40% (40/100); when NLI predicted NOT_SUPPORTED, humans agreed 75% (75/100).

Most disagreements were topical-but-not-certifying (the 'apple vs orchard' pattern), where a quote relates to the topic but does not strictly certify broader or meta claims in the justification.

Expanded human audit (n=200). Using calibrated, claimlet-level hypotheses that better match atomic receipts to the judge's justification, we audited an additional 200 receipt–justification pairs. Conditional on presence-valid receipts, binary NLI–human agreement on linkage is 87.0% (160/184) using the main-paper linkage criteria (precision 0.84, recall 0.82); see Appendix C.3.

We report certification rates as relative judge fingerprints rather than absolute truth; the pilot suggests thresholds can be tuned to trade recall for higher precision if certification is used as a gating signal.

**Shotgun Index:** Computed as total_receipts × (1 - linkage_rate). High values indicate "evidence spam" (judges who cite many passages without semantic grounding).

### 4.4 Canary Checks

To ensure attribution results are not artifacts, we run: (1) permutation tests with shuffled labels (300 iterations), (2) leave-one-video-out (LOVO) cross-validation ensuring zero group overlap, (3) tokens-only proxy probes to rule out length-based shortcuts, and (4) class-balance audits confirming 360 rows per judge in the YouTube study. All LOVO splits are audited to ensure exactly one held-out video per fold.

## 5. Results

### 5.1 Between-Judge Agreement is Low and Heterogeneous

Across the 36 judge pairs, Spearman ρ has mean 0.282 (median 0.266), ranging from 0.004 to 0.586. Krippendorff's α is 0.042 overall, indicating near-zero absolute agreement.

**The overall α of 0.04 is well below acceptable thresholds.** Convention treats α < 0.67 as inadequate for drawing conclusions; α < 0.20 indicates poor agreement. We're at 0.04.

Krippendorff's α by dimension:

| Dimension | α (interval) |
|---|---|
| Intent & Angle | 0.050 |
| Coverage & Completeness | 0.132 |
| Faithfulness & Receipts | 0.090 |
| Readability & Structure | -0.064 |
| SEO Mechanics | -0.047 |

Negative alphas indicate systematic disagreement (worse than chance) on these dimensions. On Readability and SEO Mechanics in particular, judges appear to apply incompatible criteria, so high scores from one judge do not reliably predict high scores from another.

### 5.2 Individual Judges Are Stable With Themselves

Despite low inter-judge agreement, many judges are self-consistent across the three runs. ICC(3,1) spans -0.038 to 0.872 across judges:

| Judge | ICC(3,1) |
|---|---|
| Gemini-3-Pro | 0.872 |
| GPT-5.2 | 0.845 |
| Claude-Opus | 0.811 |
| Mistral-Large | 0.758 |
| Grok-3 | 0.537 |

| Judge | ICC(3,1) |
|---|---|
| Claude-Sonnet | 0.499 |
| DeepSeek-R1 | 0.329 |
| GPT-4.1 | 0.320 |
| Llama-405B | -0.038 |

Six judges have ICC > 0.5; three exceed 0.8. These judges aren't noisy; they're stable. They just happen to be stable in different directions.

**The puzzle crystallizes:** If judges were noisy measurements of a shared truth, low inter-judge agreement would imply low intra-judge consistency. But we see the opposite: judges are stable with themselves while disagreeing with each other. This can only mean one thing: they're measuring different things.

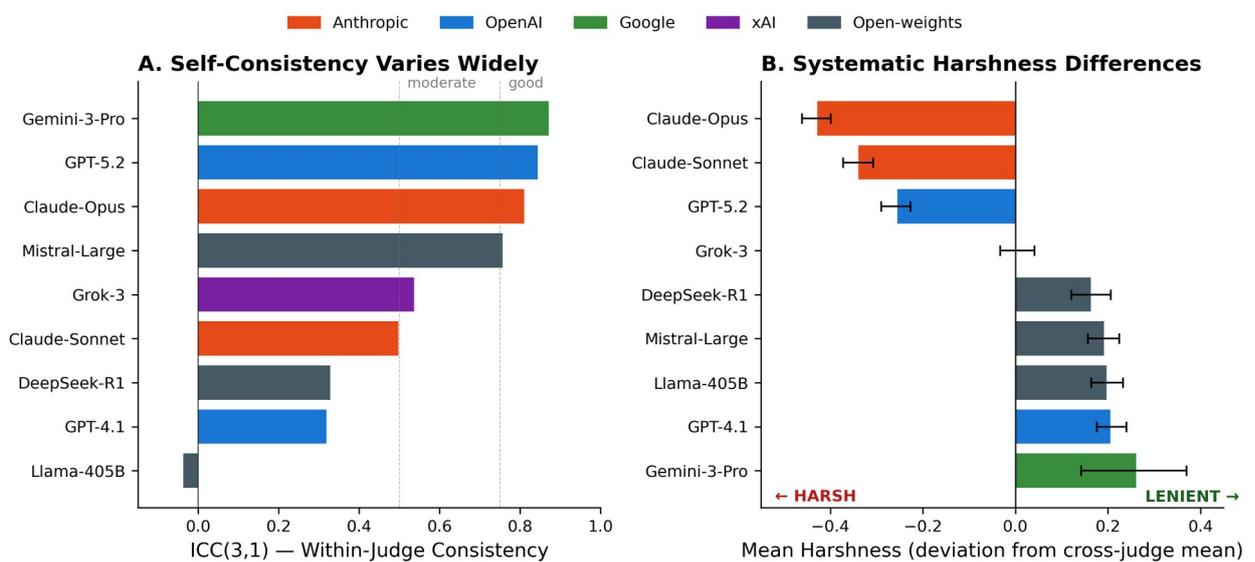

*Figure 1. The reliability paradox. (A) Within-judge consistency (ICC) varies from -0.04 to 0.87; judges are highly stable with themselves. (B) Systematic harshness differences with 95% bootstrap CIs. Claude models are consistently harsh; Gemini is consistently lenient. Non-overlapping intervals confirm real dispositional differences.*

### 5.3 Judges Differ Systematically in Harshness

Harshness estimates show consistent leniency differences. Intervals are video-cluster bootstrap 95% percentile CIs:

| Judge | Mean Harshness | 95% CI |
|---|---|---|
| Claude-Opus | -0.429 | [-0.461, -0.399] |
| Claude-Sonnet | -0.340 | [-0.372, -0.307] |
| GPT-5.2 | -0.256 | [-0.290, -0.227] |
| Grok-3 | +0.003 | [-0.033, +0.041] |
| DeepSeek-R1 | +0.164 | [+0.121, +0.206] |
| Mistral-Large | +0.192 | [+0.157, +0.224] |
| Llama-405B | +0.198 | [+0.164, +0.233] |
| GPT-4.1 | +0.206 | [+0.176, +0.240] |
| Gemini-3-Pro | +0.262 | [+0.142, +0.370] |

Claude models are systematically harsh (negative); Gemini is systematically lenient (positive). The CIs don't overlap; these are real dispositional differences.

**But harshness isn't the whole story.** Each judge also has a distinctive shape of harshness across dimensions. GPT-5.2 is uniquely harsh on Faithfulness (-0.64 deviation) while being moderate elsewhere. Claude models are uniformly harsh across all dimensions. Each judge has a recognizable "signature."

**5.4 Evidence Behavior Varies by Judge**

Across all judges we observe 50,436 raw receipts across all five rubric dimensions. For the evidence-behavior analyses below (provenance and NLI linkage), we focus on the 31,232 receipts attached to the three content-grounding dimensions (Intent, Coverage, Faithfulness), where citation-to-justification tests are meaningful.

**Provenance (Presence Validity)**

Overall validity in this subset is 94.9% (29,629/31,232), but ranges from 80.3% (Llama-405B) to 98.5% (Claude-Opus). Llama-405B cites content that isn't in the source about 19% of the time, a distinct failure mode.

| Judge | Presence-valid rate (I/C/F) | Receipts/Eval (I/C/F) |
|---|---|---|
| Claude-Opus | 98.5% | 10.7 |
| GPT-5.2 | 98.4% | 11.0 |
| Grok-3 | 97.4% | 9.2 |
| Claude-Sonnet | 96.8% | 11.8 |
| GPT-4.1 | 96.4% | 7.6 |
| Mistral-Large | 94.1% | 11.5 |
| DeepSeek-R1 | 93.8% | 9.7 |
| Gemini-3-Pro | 92.0% | 9.3 |
| Llama-405B | 80.3% | 6.1 |

**Semantic Linkage (NLI)**

Among presence-valid receipts, semantic linkage rates vary about 3× across judges, ranging from 15.4% to 44.2%. This measures whether the cited evidence actually anchors the judge's reasoning under our strict linkage test:

| Judge | NLI linkage rate (given presence-valid) | Mean margin |
|---|---|---|
| GPT-4.1 | 43.6% | +0.409 |
| Mistral-Large | 44.2% | +0.409 |
| Grok-3 | 39.7% | +0.384 |
| Claude-Opus | 25.4% | +0.247 |
| Claude-Sonnet | 30.6% | +0.298 |
| Gemini-3-Pro | 17.7% | +0.158 |
| GPT-5.2 | 37.1% | +0.354 |
| DeepSeek-R1 | 15.4% | +0.097 |
| Llama-405B | 25.9% | +0.190 |

GPT-4.1 is a "careful citer": high semantic linkage (43.6%) with comparatively low citation volume (7.6 receipts/eval). It cites less, but more often produces quotes that strictly certify its justification.

**Shotgun Index (Evidence Spam)**

Shotgun index = total_receipts × (1 - linkage_rate). High values indicate judges who spray citations without semantic grounding:

| Judge | Shotgun Index | Interpretation |
|---|---|---|
| Claude-Sonnet | 8.2 | High volume, moderate grounding |
| GPT-5.2 | 6.9 | Moderate volume, moderate grounding |
| Mistral-Large | 6.4 | High volume, high grounding |
| Claude-Opus | 8.0 | Moderate volume, low grounding |
| DeepSeek-R1 | 8.2 | Moderate volume, low grounding |
| Gemini-3-Pro | 7.6 | Moderate volume, low grounding |
| Grok-3 | 5.5 | Moderate volume, moderate grounding |
| GPT-4.1 | 4.3 | Low volume, high grounding |
| Llama-405B | 4.5 | Low volume, low validity |

**Evidence Behavior as Disposition Axis**

Combining harshness with evidence behavior reveals distinct judge "personalities":

- **GPT-4.1 (Lenient + Careful):** Scores generously (+0.21), cites sparingly, grounds well (43.6% linkage)

- **Claude-Opus (Harsh + Moderate):** Scores strictly (-0.43), cites heavily, moderate grounding (25.4% linkage)

- **Llama-405B (Lenient + Sloppy):** Scores generously (+0.20), low validity (80%), mixed grounding (25.9% linkage)

- Claude-Sonnet (Harsh + Shotgun): Scores strictly (-0.34), highest citation volume, moderate grounding (30.6% linkage)

This is not just "different calibration"; it's fundamentally different evaluation strategies.

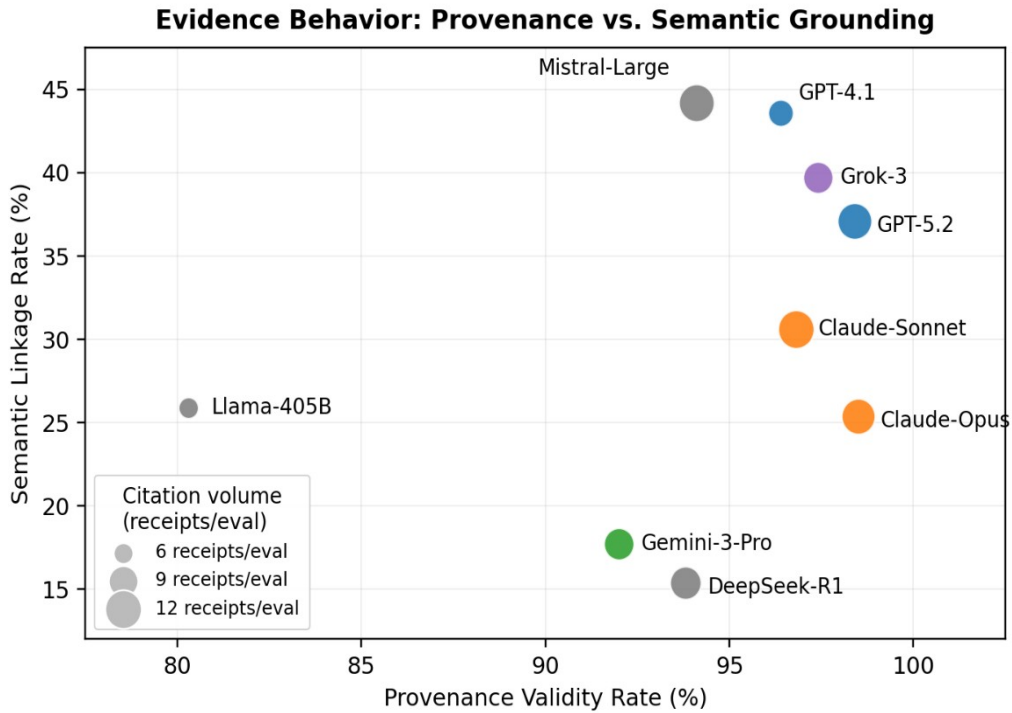

*Figure 2. Evidence behavior reveals distinct judge "personalities." Bubble size indicates citation volume. GPT-4.1 achieves high validity and semantic grounding with sparse citations. Llama-405B has low validity; 20% of its citations reference non-existent content.*

Judges also differ in which source they preferentially cite when producing receipts (pack vs. script), a stable provenance-selection trait reported in Appendix E.

**5.5 Judges Are Fingerprintable From Evaluation Behavior**

We treat judge attribution as a discriminative probe: can we predict which judge produced an evaluation row, using only the structured outputs?

**Grouped Cross-Validation (Primary Result)**

Using StratifiedGroupKFold (5 folds, grouped by video_id), we test multiple feature sets:

Behavioral compliance is intentionally part of the fingerprint. The disposition feature set is derived from the receipt-audit pipeline and includes per-dimension receipt volume, presence-validity rates, NLI linkage rates and entailment margins, shotgun indices, and pack-vs.-script splits (the disp_* metrics), along with explicit indicators for undefined ratios (e.g., zero-denominator cases when a judge produces no receipts for a slice) before filling those NaNs with 0.0 for modeling. We featurize missingness rather than hiding it via imputation, because a judge's rate and type of protocol/evidence failures is itself a stable disposition trait.

| Task | Feature Set | Accuracy | vs. Chance |
|---|---|---|---|
| Exact judge (9-way) | Scores only (5 dims) | 77.1% | 6.9× |
| Exact judge (9-way) | Disposition only | 71.5% | 6.4× |
| Exact judge (9-way) | Scores + Disposition | 89.9% | 8.1× |
| Provider lineage (7-way) | Scores + Disposition | 91.5% | 6.4× |
| Within Claude (2-way) | Scores only | 83.5% | 1.7× |
| Within Claude (2-way) | Scores + Disposition | 91.2% | 1.8× |
| Within GPT (2-way) | Scores only | 97.8% | 2.0× |
| Within GPT (2-way) | Scores + Disposition | 99.6% | 2.0× |

**The within-family results are striking.** GPT-4.1 and GPT-5.2 (models from the same provider, different versions) are distinguishable with 99.6% accuracy. Their evaluation behaviors differ enough to identify them nearly perfectly.

**Conservative LOVO Validation**

Leave-one-video-out (30 folds) provides a stricter lower bound, ensuring the classifier cannot exploit video-specific patterns:

| Feature Set | LOVO Accuracy | Macro F1 | p-value |
|---|---|---|---|
| Scores only | 37.4% | 0.331 | 0.003 |
| Combined | 59.8% | 0.564 | 0.003 |

LOVO accuracy (37-60%) is lower but still 3-5× chance, confirming generalization to unseen content types. Permutation tests with shuffled labels yield accuracy near chance (8.2%), confirming the signal is real.

**It's Not Just Global Harshness**

We test whether we're just detecting "who's harsh" by row-demeaning, subtracting each evaluation's mean score, leaving only the shape:

| Feature Set | Accuracy |
|---|---|
| Scores only | 77.1% |
| Shape only (row-demeaned) | 62.5% |
| Disposition only | 71.5% |

Shape-only accuracy (62.5%) is still 5.6× chance. Even after removing global harshness, the pattern of how judges score across dimensions is identifiable. The fingerprint isn't just "harsh vs. lenient"; it's the full dispositional profile.

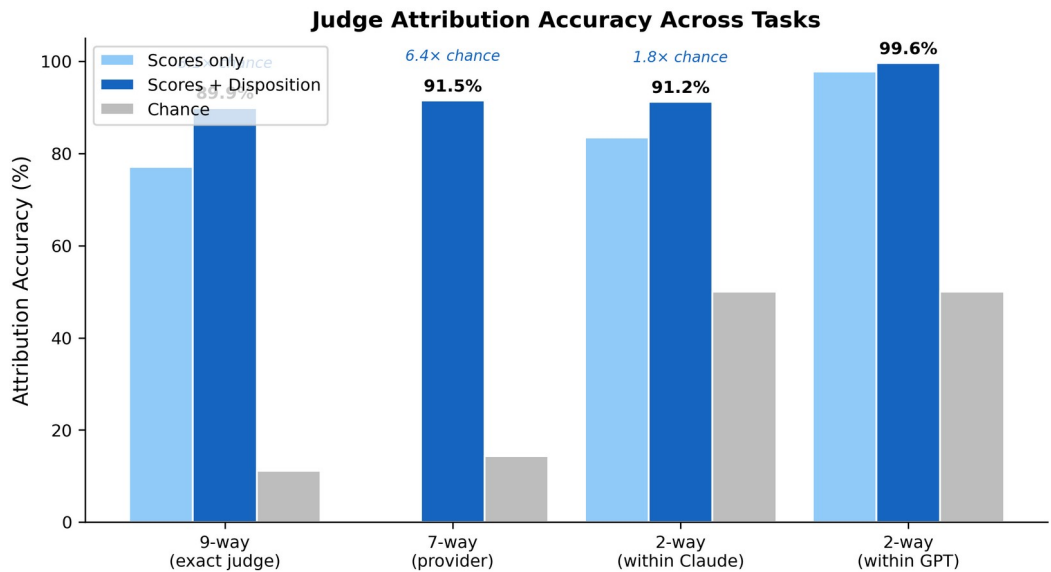

*Figure 3. Attribution accuracy across classification tasks (YouTube study). Combined features achieve 89.9% on 9-way exact judge identification (8.1× chance). Within-GPT discrimination reaches 99.6%.*

Appendix F reports an additional oracle-conditioned control that strips per-judge scoring marginals via z-score and quantile normalization; fingerprints persist after this marginal stripping.

## 6. Robustness Checks

### 6.1 Perturbation Stability

We test whether fingerprints survive prompt perturbation on an 8-video subset. Comparing per-judge mean harshness on the matched 8-video slice of the main set versus the perturbed evaluations, harshness ordering is nearly identical (Pearson r = 0.990). The extremes are preserved: Claude-Opus remains harshest and Gemini remains most lenient. On the perturbed set itself (n = 844 evaluations), grouped-CV judge attribution remains high: 69.9% accuracy from rubric scores alone, rising to 86.0% with the disposition feature set.

### 6.2 Temperature Sensitivity

On a 5-video subset, we vary temperature (0.0, 0.3, 0.7) for DeepSeek-R1 and Gemini-3-Pro. For each of 10 (model, video) combinations, we tested whether mean overall score differed across temperatures (one-way ANOVA), applying Bonferroni correction across the 10 tests (adjusted α = 0.005). Only 1/10 tests was significant after correction (DeepSeek, Video 3, $p < 0.001$). Effect sizes were small ($\eta^2 < 0.10$). Mean score ranges across temperatures were 0.047 points for Gemini and 0.133 points for DeepSeek, while the inter-model dispositional gap averaged 0.27 points (range: 0.18–0.41).

**6.3 Canary Check Summary**

Our audit bundle confirms the attribution signal is not driven by artifacts:

- **Shuffled-label test:** Accuracy drops to ~8% (chance) when judge labels are permuted
- **Tokens-only probe:** Using only token counts yields near-chance accuracy
- **LOVO fold audit:** Exactly one held-out video per fold, zero group overlap
- Class balance: 360 evaluation instances per judge (balanced; YouTube study)

**6.4 Cross-Domain Validation**

To test whether fingerprints generalize beyond the YouTube/SEO domain, we conducted a second-regime study using Wikipedia articles as source material and a different artifact type.

**Study Design.** We evaluated 15 Wikipedia articles across diverse topics (Photosynthesis, Quantum Mechanics, Roman Empire, etc.). Instead of SEO content packs, judges evaluated Structured Briefing Packs, a format-heavy template with 8 required sections (TL;DR, Key Takeaways, Core Explanation, Key Entities, Timeline, Common Misconceptions, FAQ, and Glossary). The rubric retained the same 5 dimensions with identical scoring anchors, renaming only "SEO Mechanics" to "Task Mechanics" to reflect format compliance.

**Controlled Variants.** For each article, we generated 4 artifact variants: (1) Clean: faithful, complete, well-structured; (2) Hallucination-poisoned: 3-5 injected false claims; (3) Coverage-poisoned: faithful but missing 40-50% of key subtopics; (4) Structure-poisoned: deliberate format violations. Each variant targets a specific dimension, enabling diagnostic analysis of judge capabilities.

All Wikipedia briefing-pack variants were generated by a single generator model (GPT-4.1), holding generator identity constant; thus this regime also controls for generator-judge confounds: evaluative fingerprints persist even when generator identity is held constant.

Human manipulation check. To verify that the Wikipedia controlled variants truly instantiate the intended perturbations, we sampled 40 briefing packs per condition and asked three human reviewers to label each pack by condition (Clean, Hallucination, Coverage, or Structure). Detection rates were 92.5% (Clean), 87.5% (Hallucination), 95.0% (Coverage), and 82.5% (Structure).

Cross-Domain Attribution Results. With 1,066 evaluations (planned 1,080 = 15 articles × 4 variants × 9 judges × 2 runs; 14 excluded after parse_ok filtering), fingerprints transferred strongly:

| Feature Set | YouTube | Wikipedia |
|---|---|---|
| Scores only | 77.1% | 80.9% |
| Disposition only | 71.5% | 77.0% |
| Combined (Scores + Disposition) | 89.9% | **90.3%** |
| Within-family: GPT (2-way) | 99.6% | **100%** |

The cross-domain classifier achieved 90.3% accuracy, matching the YouTube domain despite using one-third the data (1,066 vs. 3,240 evaluations). Harshness ordering was preserved: Claude models remained harshest, Mistral and Gemini remained most lenient.

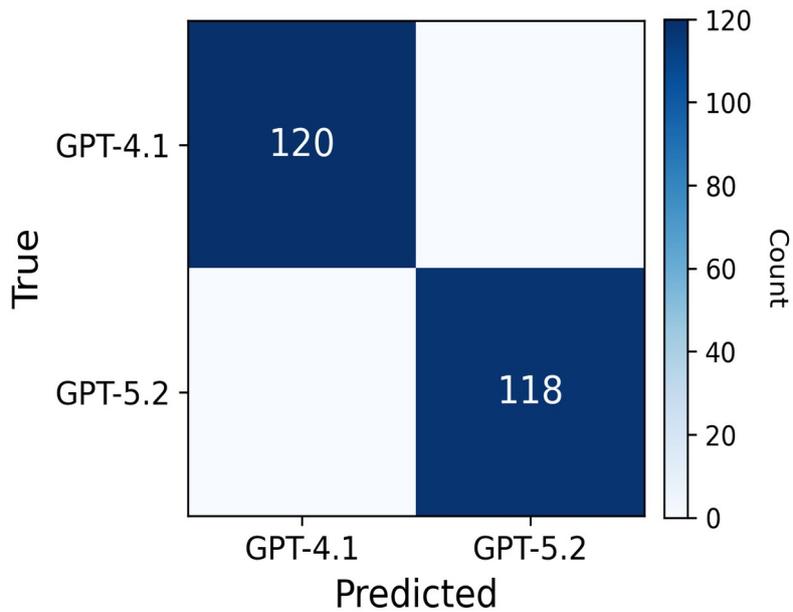

Figure 4. Within-GPT discrimination achieves 100% accuracy in cross-domain validation. Zero off-diagonal entries; GPT-4.1 and GPT-5.2 are perfectly distinguishable from evaluation behavior alone.

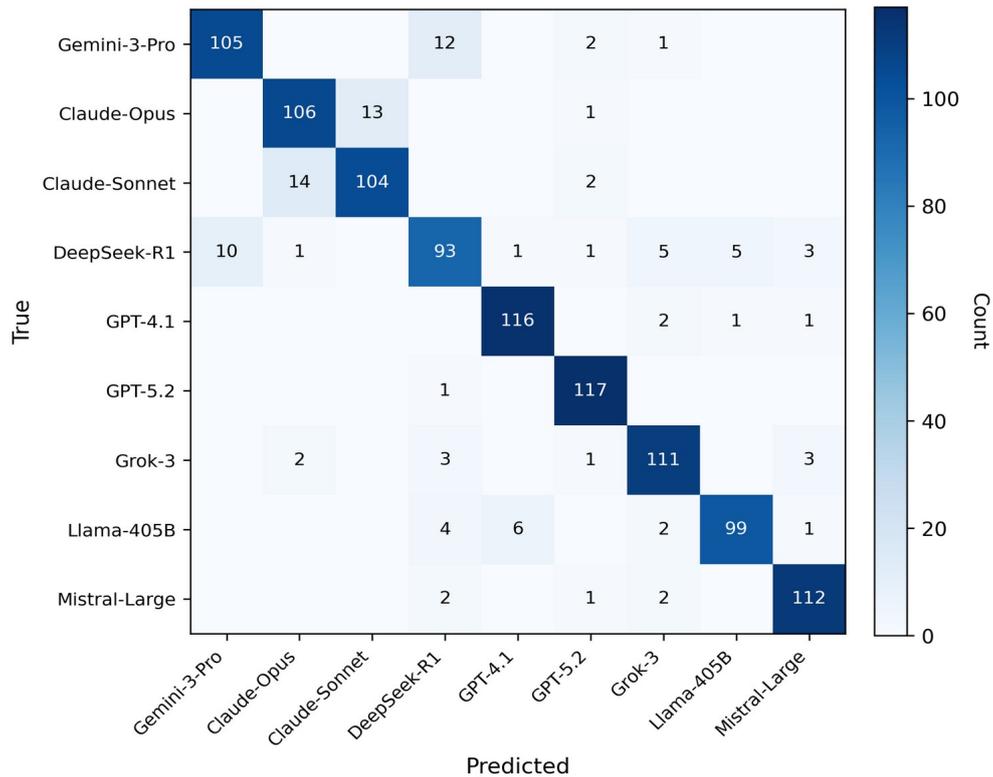

Figure 5. Confusion matrix for cross-domain validation (Wikipedia study, n=1,066). Combined features achieve 90.3% accuracy, matching YouTube performance. Strong diagonal indicates reliable identification across domains.

Hallucination Detection as Capability Fingerprint. The controlled variants revealed that fingerprints predict real capability differences. We measured each judge's faithfulness score drop between clean (pack 11) and hallucination-poisoned (pack 22) variants:

| Judge | Clean | Hallucinated | Drop | Verdict |
|---|---|---|---|---|
| Gemini-3-Pro | 4.73 | 3.27 | **-1.46** | Catches |
| GPT-5.2 | 4.34 | 3.22 | **-1.12** | Catches |
| Claude-Sonnet | 4.13 | 3.21 | **-0.92** | Catches |
| DeepSeek-R1 | 4.51 | 3.60 | **-0.91** | Catches |
| Claude-Opus | 4.08 | 3.30 | **-0.78** | Catches |
| GPT-4.1 | 4.41 | 4.09 | -0.32 | Weak |
| Grok-3 | 4.15 | 3.92 | -0.23 | Weak |
| **Mistral-Large** | 4.28 | 4.29 | **+0.01** | **BLIND** |
| **Llama-405B** | 4.12 | 4.39 | **+0.27** | **BLIND** |

Mistral-Large and Llama-405B (the same judges that were lenient on YouTube and had low evidence linkage rates) are blind to hallucinations. They rate fabricated content as equally or more faithful than clean content. The effect is even starker when examining failure rates: Gemini-3-Pro assigned faithfulness scores ≤3 to 60% of hallucinated variants, while Mistral-Large, Llama-405B, and Grok-3 never assigned a failing score (0%). This demonstrates that evaluative fingerprints are not merely stylistic: they predict genuine capability differences with practical consequences.

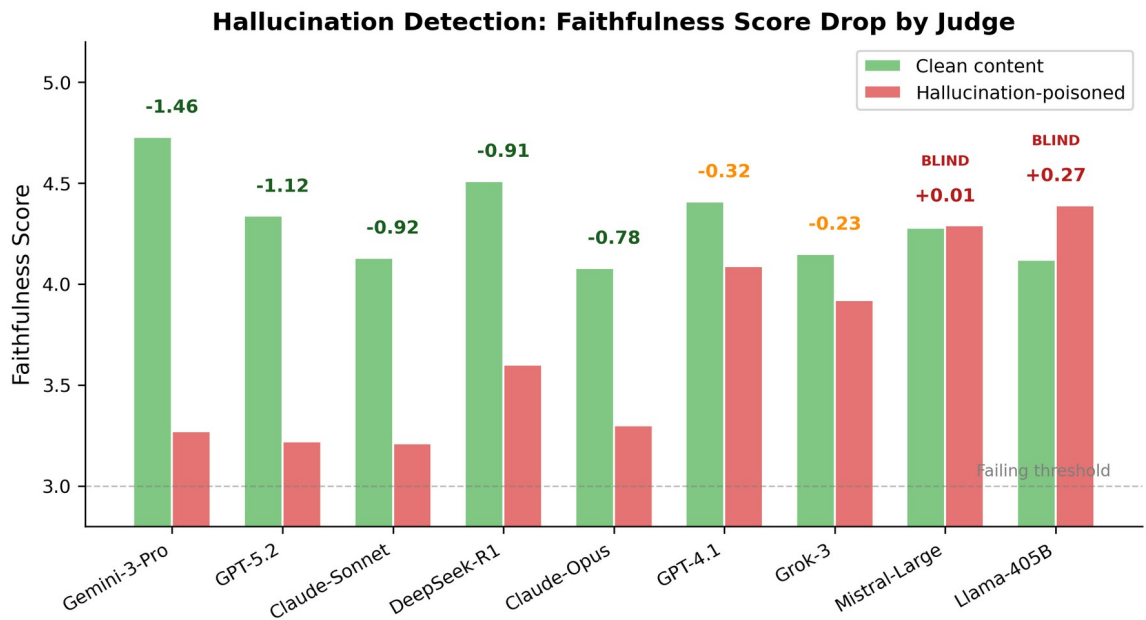

Figure 6. Hallucination detection capability varies dramatically. Effective judges show large faithfulness score drops on poisoned content. Mistral-Large and Llama-405B are "blind"; they never assign failing scores to fabricated content.

## 7. Discussion

### 7.1 The Core Finding

**Judges can't agree on what's good (α = 0.04), yet they're so consistent in how they disagree that you can identify them (89.9% accuracy).**

This is only possible if each judge embeds a distinct, stable theory of quality: an "evaluative disposition" that shapes how it interprets the rubric. These dispositions are stable (high within-judge ICC), distinctive (77-91% attribution accuracy), and hierarchical (detectable at family, model, and version levels).

### 7.2 Practical Implications

**For benchmark designers:** Your choice of judge is a methodological decision. Different judges may rank systems differently. Report which judge you used; consider reporting multiple.

**For ensemble methods:** Averaging judges doesn't give you "ground truth." It gives you a synthetic compromise that matches no judge's actual values.

**For RLHF practitioners:** The reward model's evaluation disposition shapes what behaviors get reinforced. Training on Claude vs. GPT feedback may produce meaningfully different models.

**For auditors:** Evaluation behavior can reveal model identity. Implications for provenance, accountability, and detecting undisclosed model changes.

### 7.3 What This Does Not Claim

We measure consistency of reasoning, not correctness. High linkage rate does not mean a judge is "right"; it means its cited evidence semantically aligns with its stated justification. A judge could be consistently wrong. Our contribution is showing that judges have stable, identifiable evaluation strategies, not that any strategy is superior.

### 8. Limitations

**Domain:** Primary study uses 30 videos with one rubric (SEO content evaluation). Cross-domain validation (Section 6.4) demonstrates fingerprint persistence on Wikipedia/briefing-pack content, but additional domains and rubrics remain untested.

**Temporal:** Three runs per judge over a short period. Long-term stability and temporal drift remain open questions.

**Evidence validation:** We report both provenance (presence) and semantic linkage (NLI); this does not establish ground truth, and NLI thresholds introduce model-based measurement error.

Judge sample: 9 models from 5 high-level provider families (and 7 provider lineages when open-weights are split). Different panels may show different patterns.

**Feature definitions:** "Scores-only" means the 5 rubric dimension scores; results may vary with different feature engineering choices.

### 9. Reproducibility

Code and artifacts: All reproducibility materials referenced in this paper are hosted in a companion repository, Evaluative Fingerprints[2]. We also release (i) the evaluation harness used to run all judge calls for this study (12 Angry Tokens[1], a multi-judge LLM evaluation harness); (ii) the exact judge prompt and scoring rubric; (iii) analysis code reproducing all tables and figures; and (iv) complete parsed judge JSON outputs for the 30-video study. Reproduction does not require rerunning the harness; Evaluative Fingerprints[2] includes self-contained scripts for all analyses.

[1] https://github.com/Wajid-Nasser/12-Angry-Tokens
[2] https://github.com/wajid-nasser/evaluative-fingerprints

Copyright constraints: We do not redistribute full transcripts or SEO packs. We provide a 3-video open sample (rights-cleared / non-copyrighted) for end-to-end verification.

Reproduction modes: (a) Full pipeline reproduction on the licensed sample, or (b) statistic reproduction on released judge JSON artifacts.

Supplementary artifact bundle (no API keys, no GPU): In Evaluative Fingerprints², we include a self-contained directory with fully precomputed outputs for the three primary result sets (30-video study, 8-video perturbation subset, and 15-article Wikipedia second-regime study). Each dataset folder contains a single authoritative output directory named `run_complete_pipeline_<timestamp>` with the final tables, figures, consolidated CSVs, and audit report. Reproduction is scriptable and does not rerun NLI, classifiers, or any model calls.

- `30_vid_set/` (main 30-video YouTube study)
- `vid_8_perturbed_set/` (8-video perturbation robustness subset)
- `wiki_15_set/` (15-article Wikipedia second-regime study)

Scripts at the bundle root copy the corresponding `run_complete_pipeline_*` directory into `reproduced_results/` for easy inspection (Windows: `reproduce_all_primary_sets.ps1`; Linux/macOS: `reproduce_all_primary_sets.sh`).

## 10. Conclusion

We set out to test whether LLM judges are interchangeable instruments. They are not.

Judges agree poorly with each other ($\alpha = 0.04$) but consistently with themselves (ICC up to 0.87). Their disagreement patterns are structured enough to identify which model produced an evaluation with 89.9% accuracy. Even models from the same provider (GPT-4.1 vs. GPT-5.2) are distinguishable with 99.6% accuracy.

The reliability paradox (low agreement, high identifiability) reveals that LLM-as-judge is not one thing. Each judge implements a different theory of quality. Using one means adopting its values. Averaging them means adopting no one's values.

The path forward isn't to find the "right" judge. It's to understand what each judge measures, report results transparently, and treat model selection as the methodological choice it is.

## References

Zheng, L. et al. (2023). Judging LLM-as-a-Judge with MT-Bench and Chatbot Arena. NeurIPS.

Evaluative Fingerprints

**Appendix A: Perturbation Subset Details**

We conducted a robustness check on an 8-video perturbation subset. Videos were selected to span topic diversity (comedy, tech, travel, sports).

Perturbations included: minor formatting changes, whitespace normalization, and content-preserving synonym substitutions in the SEO packs.

Results: Harshness ordering correlation r = 0.990 (Pearson) between the matched unperturbed 8-video slice and the perturbed evaluations (computed on per-judge mean harshness). Attribution accuracy on the perturbed set (n = 844 evaluations) remained high: 69.9% from rubric scores alone and 86.0% from scores + disposition. Within-judge stability estimates on the smaller perturbed set are noisier: ICC(3,1) ranges from 0.08 to 0.75 across judges, with n_items = 26–32 depending on missing item–run triples.

Conclusion: Judge fingerprints are robust to surface-level input perturbations.

**Appendix B: Temperature Sensitivity Study**

Temperature grid: 0.0, 0.3, 0.7 for DeepSeek-R1 and Gemini-3-Pro (5-video subset).

For each of 10 (model, video) combinations (2 models × 5 videos), we tested whether mean overall score differed across temperatures using a one-way ANOVA across the three temperatures, then applied Bonferroni correction across the 10 tests (adjusted α = 0.005). Only 1/10 tests was significant after correction (DeepSeek, Video 3, p < 0.001).

Interpretation: Temperature settings induced only small score variation (mean range 0.047 points for Gemini; 0.133 points for DeepSeek) relative to between-model dispositional differences (mean inter-model gap 0.27 points). This supports evaluative disposition as a stable model characteristic rather than a temperature artifact.

**Appendix C: Receipt Validation Methodology**

**C.1 Provenance (Presence) Validation**

Receipt text is normalized (lowercased, whitespace-collapsed, punctuation-stripped) and matched against the source document using RapidFuzz with threshold 0.90.

Match type distribution across the 31,232 analyzed receipts (Intent, Coverage, Faithfulness): 77.1% exact, 17.8% fuzzy matches (0.90–0.99 similarity), and 5.1% no match (presence-invalid).

**C.2 Semantic Linkage (NLI) Validation**

For presence-valid receipts with non-empty justifications, we run NLI using a DeBERTa-v3 NLI model fine-tuned on MNLI.

Premise: quoted receipt text. Hypothesis: truncated justification (first 200 characters).

Linkage criteria: p_entailment ≥ 0.75 AND margin (p_entail - p_contradict) ≥ 0.20.

Overall linkage rate: 31.1% of presence-valid receipts meet linkage criteria.

## C.3 Calibrated human audit for Semantic Linkage

We performed a second, claimlet-level audit to reduce granularity mismatch between aggregated justifications and atomic receipts. We sampled 200 receipt–justification pairs and collected human labels (ENTAILMENT vs NEUTRAL; NONE for presence-invalid). We then evaluated binary linkage predictions using the main-paper linkage criteria (p_entailment ≥ 0.75 and margin ≥ 0.20).

| Metric | Value |
| --- | --- |
| n (total) | 200 |
| n (presence-valid) | 184 |
| Binary agreement (presence-valid) | 87.0% (160/184) |
| Precision | 0.84 |
| Recall | 0.82 |
| False positives | 11 |
| False negatives | 13 |

## C.4 Exact-match sensitivity

To quantify how often receipts are copied verbatim versus lightly edited, we also compute an exact-match-only variant of presence validation (threshold = 1.0), which treats all fuzzy matches as invalid.

Results: Overall exact-match rate is 77.1% (24,067/31,232), compared to 94.9% with fuzzy matching (threshold = 0.90).

Conclusion: Exact matching is a conservative lower bound because it penalizes trivial formatting differences and transcript punctuation variance; we therefore treat fuzzy matching as the primary operationalization and report the exact/fuzzy split in Appendix C.1.

**Appendix D: Per-Generator Consistency**

We tested whether fingerprints hold across different content generators (GPT-5.2, GPT-4.1, Gemini-3-Pro, Mistral/Claude packs).

In this appendix, pack IDs identify the generator LLM used to create each pack: 3 = Gemini 3 Pro; 41 = GPT-4.1; 45 = Claude Opus; 53 = Mistral v3 Large; 523 = GPT-5.2-chat.

Per-pack attribution accuracy (scores + disposition): Pack 3: 87.3%, Pack 41: 87.8%, Pack 45: 80.8%, Pack 53: 83.6%, Pack 523: 86.5%.

Conclusion: Judge fingerprints generalize across generator sources, not just video topics.

# Appendix E: Receipt Source Preference (Pack vs. Script)

## E.1 Definition

We define receipt source preference as the fraction of extracted receipts that are attributed to the SEO pack versus the script (the two primary evidence substrates compared by the judge). Each receipt is classified as PACK or SCRIPT based on the receipt's declared source field; a small residual category OTHER captures receipts that do not map cleanly to either substrate under our parser.

## E.2 Results

Table E.1 reports receipt source preference on the primary 30-video YouTube study. These rates describe citation selection behavior (which source is cited), not scoring quality or factual correctness.

| Judge | N evals | Pack receipts | Script receipts | Other | Pack rate (%) |
|---|---|---|---|---|---|
| GPT-4.1 | 360 | 3874 | 619 | 0 | 86.22 |
| GPT-5.2 | 360 | 5217 | 1264 | 0 | 80.50 |
| Grok-3 | 360 | 4270 | 1157 | 0 | 78.68 |
| Mistral-Large | 360 | 5192 | 1755 | 0 | 74.74 |
| DeepSeek-R1 | 360 | 4246 | 1433 | 0 | 74.77 |
| Claude-Opus-4.5 | 360 | 4312 | 1723 | 0 | 71.45 |
| Llama-405B | 360 | 2564 | 1051 | 6 | 70.81 |
| Claude-Sonnet-4.5 | 360 | 4915 | 2019 | 0 | 70.88 |
| Gemini-3-Pro-Preview | 360 | 3395 | 1424 | 0 | 70.45 |

## E.3 Interpretation and Caveats

- Receipt source preference reflects provenance-selection behavior under our current evidence presentation and parsing; it should not be interpreted as a correctness metric.

- Rates may be partially confounded by rubric dimension (e.g., readability critiques may naturally cite pack text), and by differences in how quoteable each substrate is (span length, formatting, and salience).

- Request retries in our pipeline are dominated by malformed JSON outputs or provider rate limiting; retries are not triggered by receipt quality, so the table captures selection rather than verified grounding.

# Appendix F: Oracle Normalization Controls (Per-Judge Marginal Stripping)

## F.1 Motivation

We test whether judge attribution can be explained solely by per-judge scoring scale usage (e.g., persistent leniency/harshness or dimension-wise calibration). As an oracle-conditioned control, we normalize rubric scores within each judge (using parameters fit on the training fold only) and rerun the attribution probe.

## F.2 Setup

Data: 30-video YouTube study intersection set (3,240 evaluations; 9 judges; 120 unique video×pack items; 3 runs).

Features: 5 rubric dimension scores.

Cross-validation: StratifiedGroupKFold (5 folds), grouped by video_id.

Classifier: Random forest (rf).

Normalizations (fit on the training fold; applied to train and test within each fold):

Per-judge z-score: for each judge and dimension, subtract the judge mean and divide by the judge standard deviation.

Per-judge quantile: for each judge and dimension, map each score to its within-judge empirical CDF rank in [0, 1].

## F.3 Results

| Normalization | Accuracy | Macro F1 |
| --- | --- | --- |
| Per-judge z-score (train-fold fit) | 0.9790 | 0.9790 |
| Per-judge quantile rank (train-fold fit) | 0.9769 | 0.9769 |

## F.4 Interpretation and Caveat

Attribution remains high after stripping per-judge marginals, indicating that fingerprints are not reducible to global harshness/leniency or simple score-scale usage. Because the marginal-stripping transformation conditions on judge identity, this is not a deployable preprocessing step; it is a control analysis intended to isolate dependence structure across rubric dimensions.